\documentclass[runningheads]{llncs}
\usepackage[T1]{fontenc}
\usepackage{graphicx}
\usepackage{booktabs}
\usepackage{amsmath}
\usepackage{amssymb}
\usepackage{caption}
\usepackage{bbding}
\begin{document}
\title{CDGC-Net: 3D Medical Image Segmentation with Cooperative Dual-Scale Self-Attention and Grouped Channel Modeling}
\titlerunning{CDGC-Net for 3D Medical Segmentation}
%
\author{Zheyang Jing \and
Qin Lu \and
Jianwang Li \and
Yujie Yang \and
Chen Yi \and
Shaofeng Jiang\Envelope}
\authorrunning{Z. Jing et al.}
%
\institute{Nanchang Hangkong University, Nanchang, China\\
\email{1255813468@qq.com, luq0715@163.com, 2037405746@qq.com, 1796104091@qq.com, 470835297@qq.com, jsphone@163.com}}
\maketitle              
\begin{abstract}
Accurate 3D medical image segmentation requires the integration of long-range anatomical context with fine boundary detail. Existing methods often model global and local features in separate modules or feature levels and perform channel recalibration independently. This may cause semantic mismatch between global context and local boundaries, insufficient channel relationship modeling, weak spatial-channel interaction, and redundant representations. We propose CDGC-Net, a 3D medical image segmentation network that combines cooperative dual-scale spatial attention with grouped hierarchical channel modeling. With-in each CDGC block, Cooperative Dual-Scale Self-Attention (CDSA) assigns attention heads to parallel local-window and global-sparse branches. The two branches capture fine spatial details and long-range anatomical context at the same feature level. Their outputs are concatenated into an $N\times C$ spatial representation and directly passed to Grouped Hierarchical Channel Attention (GHCA). GHCA organizes the channels into $r$ groups and models both within-group and cross-group dependencies. CDSA and GHCA reuse a shared key projection to maintain a consistent feature reference. Residual feature alignment subsequently integrates the refined features with the original representation. On the Synapse, ACDC, BraTS, and LA datasets, CDGC-Net achieved mean DSC values of 86.96\%, 92.91\%, 82.56\%, and 93.52\%, respectively, exceeding the next-highest reported values by 0.39, 0.47, 0.17, and 0.32 percentage points. CDGC-Net contains 25.83M parameters and 28.62G FLOPs for an input size of $64\times128\times128$, reducing these quantities by 39.87\% and 40.30\%, respectively, relative to UNETR++. These results indicate a favorable trade-off between segmentation accuracy and computational complexity.
\end{abstract}
\section{Introduction}
Volumetric medical image segmentation is a core task in medical image processing. It has wide applications in tumor assessment, quantitative organ analysis, surgical navigation, and treatment response tracking \cite{RN1,RN2}. These tasks require models to generate accurate 3D segmentation masks while preserving both global anatomical context and fine boundary details. Therefore, they place high demands on spatial representation and feature discrimination. Following the success of CNN-based methods, Transformers have been widely introduced into 3D medical image segmentation. Through self-attention, they can effectively model long-range dependencies and improve segmentation performance \cite{RN1,RN7}.

However, standard self-attention has a quadratic complexity of $O(n^2)$. When processing 3D volumetric data, it brings high memory usage and training cost. To reduce the computational burden, existing methods usually adopt window attention, local self-attention, deformable attention, or CNN-Transformer hybrid designs \cite{RN23,RN24,RN25,RN26}. For example, window-based or local attention methods reduce complexity by limiting the attention range. CoTr \cite{RN26} uses deformable attention to model key spatial positions. TransUNet \cite{RN6} and UNETR++ \cite{RN15} combine local feature extraction with global context modeling to enhance feature representation. Although these methods improve the balance between long-range dependency modeling and computational efficiency, they often extract global context and local details from different modules, branches, or feature levels. These features differ in spatial resolution, receptive field, and semantic abstraction. Without sufficient cross-scale interaction and alignment, feature fusion may produce a mismatch between global semantics and local boundary details. In addition, when multiple attention branches lack explicit interaction or constraints, their responses may overlap. This can reduce feature complementarity and increase the risk of redundant representations.

Besides scale differences in spatial features, semantic representation in the channel dimension also affects segmentation performance. Existing channel attention methods, such as SE \cite{RN8}, CBAM \cite{RN9}, and ECA \cite{RN29}, mainly enhance feature representation by recalibrating channel responses. However, these methods usually generate channel weights through global statistical descriptions or local cross-channel interactions. They often do not explicitly model hierarchical relationships among different channel subspaces. For medical images, responses from background, normal tissues, and lesion regions often overlap in the channel dimension. Therefore, a single channel recalibration strategy may fail to emphasize both high-response salient regions and low-response fine-grained regions. As a result, some useful features may be weakened. 
Furthermore, many segmentation networks introduce both spatial attention and channel attention to enhance feature representation \cite{RN9,RN13,RN15,RN31}. For example, CBAM-UNet \cite{RN31} adopts a serial weighting strategy with channel attention and spatial attention. TransFuse \cite{RN13} uses the BiFusion module to fuse attention features from different branches. The EPA module in UNETR++ \cite{RN15} also improves feature representation through paired spatial and channel attention. However, existing methods usually still rely on single-scale attention for spatial modeling. They also rarely model grouped hierarchical relationships in channel modeling. Meanwhile, spatial and channel features are often fused by addition, concatenation, or convolution. Their correlation modeling and feature alignment still have room for improvement. Therefore, how to cooperatively model dual-scale spatial information and structured channel relationships in a unified module, and how to build effective alignment between spatial and channel features, remain important problems in 3D medical image segmentation.

To address these problems, we propose CDGC-Net, a cooperative dual-scale self-attention and grouped channel modeling network for 3D medical image segmentation. CDGC-Net uses the U-shaped UNETR++~\cite{RN15} backbone and replaces each computation block with the proposed CDGC block. This block integrates cooperative dual-scale spatial modeling, grouped hierarchical channel modeling, and residual feature alignment to enhance the interaction between spatial and channel features.
The main contributions of this work are summarized as follows:
(1) We design a Cooperative Dual-Scale Self-Attention (CDSA) mechanism. It combines global sparse attention and local window attention to capture coarse-scale global context and fine-scale local structures within a unified spatial attention module.
(2) We introduce a Grouped Hierarchical Channel Attention (GHCA) module. It models intra-group and inter-group channel relationships to improve channel discrimination and feature complementarity.
(3) We connect CDSA and GHCA through shared key projection, spatial-guided channel query, and residual feature alignment. This forms a continuous feature refinement process and promotes effective interaction between dual-scale spatial features and grouped channel features. Experiments on Synapse, ACDC, BraTS, and LA show that CDGC-Net achieved the highest mean DSC among the compared methods under matched evaluation protocols.

\section{Methods}
\subsection{Overall Architecture}
\begin{figure}
    \centering    
    \includegraphics[width=1\linewidth]{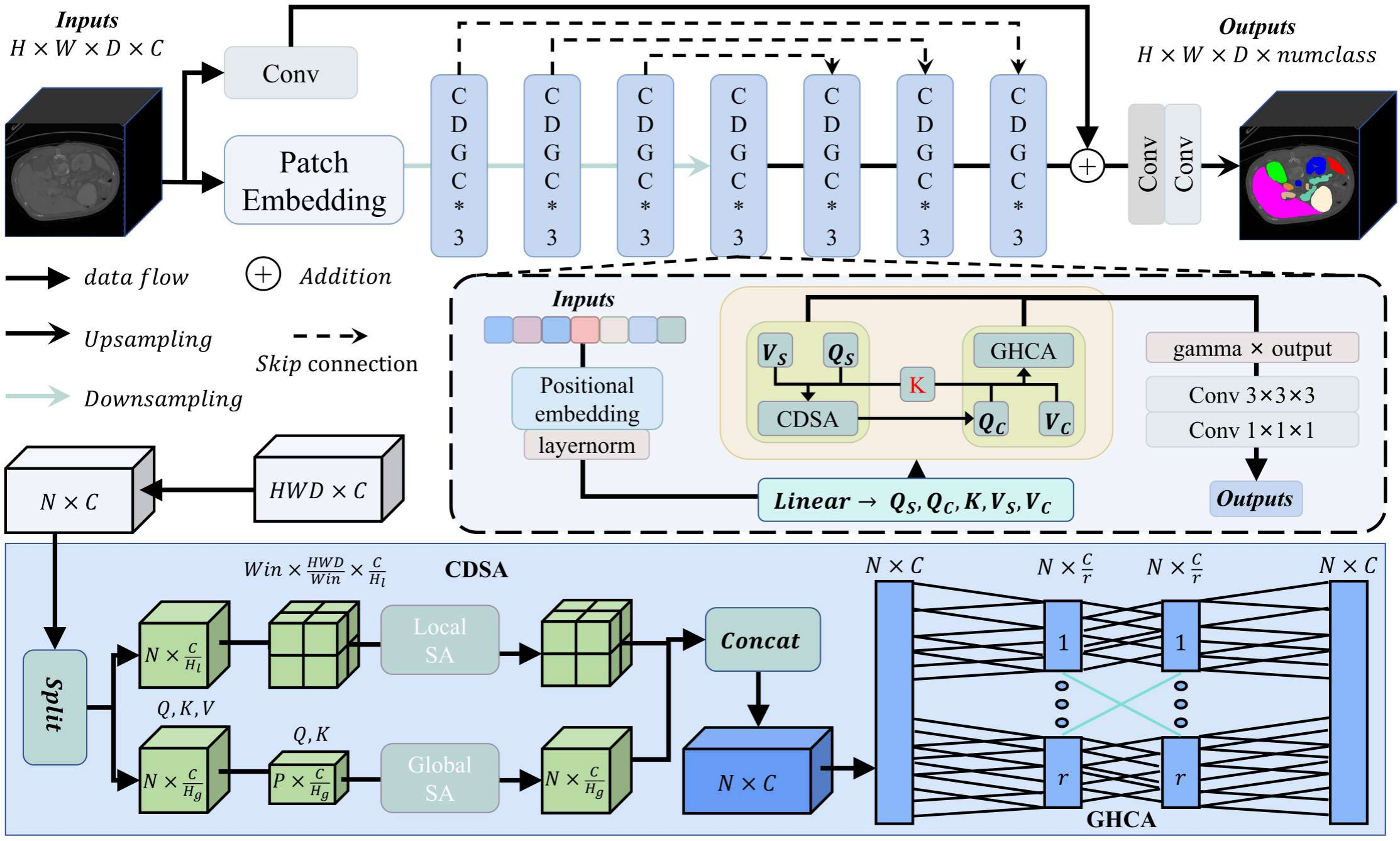}
    \caption{Overall architecture of the proposed CDGC-Net.}
    \label{fig1}
    \vspace{-3mm}
\end{figure}
As shown in Fig.~\ref{fig1}, CDGC-Net adopts the U-shaped encoder-decoder architecture of UNETR++~\cite{RN15}. Given a 3D medical image, the network first maps the input volume into embedded features through a patch embedding layer. Meanwhile, a shallow convolution branch processes the original input and preserves high-resolution spatial information for the final prediction.

During the encoding stage, the network extracts multi-scale volumetric features through four hierarchical stages. Each encoder stage stacks three CDGC blocks as the main feature extraction units. As the network goes deeper, the spatial resolution gradually decreases, while the channel dimension gradually increases. After the encoder obtains hierarchical features, the decoder gradually restores the spatial resolution through upsampling. At each decoding stage, the upsampled feature is fused with the corresponding encoder feature through a skip connection. The stacked CDGC blocks then further refine the fused feature.

Within each CDGC block, positional embedding and layer normalization produce the normalized feature $\bar{X}$. Linear projections generate the spatial query $Q_s$, shared key $K$, and spatial value $V_s$. CDSA produces the spatial representation $\hat{X}_s$, which conditions the channel query $Q_c$ in GHCA. GHCA reuses $K$ with the channel value $V_c$, after which residual feature alignment and convolutional refinement produce the block output.

After the final decoding stage, the network fuses the decoder output with the high-resolution feature from the shallow convolution branch. Finally, a prediction head with $3\times3\times3$ and $1\times1\times1$ convolutions maps the fused feature to the target classes and uses a softmax operation to generate the final segmentation probability map.

\subsection{Cooperative Dual-Scale Self-Attention (CDSA)}

CDSA captures local and global spatial dependencies at the same feature level. Given $X\in\mathbb{R}^{H\times W\times D\times C}$, CDSA first flattens the spatial dimensions to obtain $X_f\in\mathbb{R}^{N\times C}$, where $N=HWD$. After positional embedding and layer normalization, the resulting feature $\bar{X}$ is linearly projected as
\begin{equation}
Q_s=\bar{X}W_s^Q,\qquad
K=\bar{X}W^K,\qquad
V_s=\bar{X}W_s^V.
\end{equation}
The key $K$ is computed once and subsequently reused by GHCA. For CDSA, $Q_s$, $K$, and $V_s$ are partitioned into $H_l$ local heads and $H_g$ global heads. Here, $H_l+H_g=H_a$, where $H_a$ is the total number of attention heads and $d_h=C/H_a$ is the dimension of each head. The two head sets are routed to window attention and sparse global attention, respectively.

For each local head, the query, key, and value are rearranged as $Q_l,K_l,V_l\in\mathbb{R}^{\mathrm{Win}\times(N/\mathrm{Win})\times d_h}$, corresponding to $\mathrm{Win}$ non-overlapping windows with $N/\mathrm{Win}$ tokens each. Local attention is then computed as
\begin{equation}
\hat{X}_l=
\mathrm{Softmax}
\left(
\frac{Q_lK_l^T}{\sqrt{d_h}}+B
\right)V_l.
\end{equation}
Here, $d_h$ is the dimension of each attention head. For a 3D window of size $h_w\times w_w\times d_w$, let $M=h_ww_wd_w$ denote its number of tokens. The learnable relative position bias $B\in\mathbb{R}^{M\times M}$ encodes the relative spatial displacement between every pair of tokens within the window and is added independently to each local head.

For each global head, the key and value are compressed along the token dimension using learnable projection matrices $E_K,E_V\in\mathbb{R}^{P\times N}$:
\begin{equation}
\hat{K}_g=E_KK_g,\qquad
\hat{V}_g=E_VV_g,
\end{equation}
where $K_g,V_g\in\mathbb{R}^{N\times d_h}$ and $\hat{K}_g,\hat{V}_g\in\mathbb{R}^{P\times d_h}$, with $P\ll N$. Sparse global attention is then computed as:
\begin{equation}
\hat{X}_g=
\mathrm{Softmax}
\left(
\frac{Q_g\hat{K}_g^T}{\sqrt{d_h}}
\right)\hat{V}_g.
\end{equation}
The reduced key and value preserve long-range interactions while lowering the spatial attention cost.

After merging the heads, the local and global branch outputs have dimensions $\hat{X}_l\in\mathbb{R}^{N\times H_ld_h}$ and $\hat{X}_g\in\mathbb{R}^{N\times H_gd_h}$, respectively. They are then concatenated along the channel dimension:
\begin{equation}
\hat{X}_s=
\mathrm{Concat}
\left(
\hat{X}_l,\hat{X}_g
\right)
\in\mathbb{R}^{N\times C}.
\end{equation}
Thus, CDSA integrates local-window details and sparse global context without requiring features from different network levels.

\subsection{Grouped Hierarchical Channel Attention (GHCA)}

GHCA receives $\hat{X}_s\in\mathbb{R}^{N\times C}$ and models channel correlations through within-group and cross-group attention. The spatially conditioned query and channel value are defined as $Q_c=(\bar{X}+\hat{X}_s)W_c^Q$ and $V_c=\bar{X}W_c^V$, respectively. The key $K=\bar{X}W^K$ is reused from CDSA without an additional key projection.

Before channel grouping, the representations from all $H_a$ attention heads are concatenated along the channel dimension, restoring $Q_c,K,V_c\in\mathbb{R}^{N\times C}$. Each tensor is then divided into $r$ channel groups, where $Q_c^i,K^i,V_c^i\in\mathbb{R}^{N\times d_g}$, $d_g=C/r$, and $i=1,\ldots,r$. Within-group channel attention is computed as
\begin{equation}
G_i=
V_c^i
\mathrm{Softmax}
\left(
\frac{(Q_c^i)^TK^i}{\sqrt{d_g}}
\right),
\qquad i=1,\ldots,r.
\end{equation}
Here, $(Q_c^i)^TK^i\in\mathbb{R}^{d_g\times d_g}$ represents the channel affinity within the $i$-th group, and $G_i\in\mathbb{R}^{N\times d_g}$ is the corresponding group-enhanced feature. The group outputs are concatenated as $G=\mathrm{Concat}(G_1,\ldots,G_r)\in\mathbb{R}^{N\times C}$.

For cross-group modeling, $G$ is reshaped along the channel dimension into $r$ groups. Without additional linear projections, the query, key, and value are constructed as $Q_G=K_G=\mathrm{Reshape}(\mathrm{Norm}_{N}(G))$ and $V_G=\mathrm{Reshape}(G)$, where $Q_G,K_G,V_G\in\mathbb{R}^{N\times r\times d_g}$. Here, $\mathrm{Norm}_{N}(\cdot)$ denotes $\ell_2$ normalization along the token dimension. For each spatial token $n$, cross-group attention is computed as
\begin{equation}
A_G^{(n)}=
\mathrm{Softmax}\left(
\frac{Q_G^{(n)}\left(K_G^{(n)}\right)^T}{\sqrt{d_g}}
\right),
\qquad
\tilde{G}^{(n)}=A_G^{(n)}V_G^{(n)}.
\end{equation}
where $A_G^{(n)}\in\mathbb{R}^{r\times r}$ and $\tilde{G}^{(n)}\in\mathbb{R}^{r\times d_g}$. The softmax operation is applied along the last group dimension. The outputs for all spatial tokens are reshaped from $N\times r\times d_g$ back to $N\times C$, producing the channel-enhanced feature $\hat{X}_c$. The off-diagonal entries of $A_G^{(n)}$ propagate information across different channel groups. 

\subsection{Residual Feature Alignment (RFA)}
After GHCA produces $\hat{X}_c$, residual feature alignment applies a channel-wise scaling vector initialized as $\gamma^{(0)}=10^{-6}\mathbf{1}_C$.
The aligned feature is defined as
\begin{equation}
X_a=
\bar{X}+
\gamma\odot\hat{X}_c,
\end{equation}
where $\odot$ denotes channel-wise multiplication with broadcasting over the spatial tokens. This residual pathway retains the normalized input representation while controlling the magnitude of the channel correction.

The aligned feature is further refined through residual convolution:
\begin{equation}
X_{\mathrm{out}}=
X_a+
\mathrm{Conv}_{1\times1\times1}
\left(
\mathrm{Conv}_{3\times3\times3}(X_a)
\right).
\end{equation}

\subsection{Loss Function}
Following UNETR++~\cite{RN15}, UNETR~\cite{RN1}, and nnFormer~\cite{RN7}, we use a combination of soft Dice loss and cross-entropy loss for supervision. The soft Dice loss measures the overlap between the prediction and the ground truth, while the cross-entropy loss encourages voxel-wise classification accuracy. The total loss is defined as:
\begin{equation}
\mathcal{L}(Y,P) = 
1 - \frac{1}{I}\sum_{i=1}^{I} 
\frac{2\sum_{v=1}^{V}Y_{v,i}P_{v,i}+\epsilon
}{\sum_{v=1}^{V}Y_{v,i}^{2} + \sum_{v=1}^{V}P_{v,i}^{2} 
+\epsilon} - \frac{1}{V}\sum_{v=1}^{V}\sum_{i=1}^{I} Y_{v,i}\log(P_{v,i}),
\end{equation}
where $I$ denotes the number of classes and $V$ denotes the number of voxels. $Y_{v,i}$ and $P_{v,i}$ denote the ground-truth label and predicted probability for class $i$ at voxel $v$, respectively. $\epsilon$ is a small constant used to avoid numerical instability.

\section{Experiments}
\subsection{Experimental Configuration}
\textbf{Datasets.} We conduct experiments on four public 3D medical image segmentation datasets, including Synapse Multi-organ CT Segmentation~\cite{RN48}, ACDC~\cite{RN34}, BraTS~\cite{RN35}, and Left Atrium (LA) Segmentation~\cite{RN43}. 1) The Synapse dataset contains 30 abdominal CT scans with annotations of eight organs, including the spleen, right kidney, left kidney, gallbladder, liver, stomach, aorta, and pancreas. Following previous methods~\cite{RN6}, we use 18 cases for training and 12 cases for testing. 2) The ACDC dataset consists of cardiac MRI scans from 100 patients. It provides annotations for three cardiac structures: the right ventricle (RV), myocardium (MYO), and left ventricle (LV). Following nnFormer~\cite{RN7}, we split the dataset into 70 training cases, 10 validation cases, and 20 testing cases. 3) The BraTS dataset contains 484 brain tumor MRI cases. Each case includes four modalities: FLAIR, T1w, T1gd, and T2w. The segmentation targets include whole tumor (WT), enhancing tumor (ET), and tumor core (TC). We divide the dataset into training, validation, and testing sets with a ratio of 80:5:15. 4) The LA dataset contains 100 cardiac MRI scans with left atrium annotations. We use 70 cases for training, 10 cases for validation, and 20 cases for testing.

\textbf{Evaluation Metrics.} We use the Dice Similarity Coefficient (DSC) and the 95\% Hausdorff Distance (HD95) as evaluation metrics. DSC measures the overlap between the predicted segmentation and the ground truth, while HD95 evaluates boundary errors between them. A higher DSC and a lower HD95 indicate better segmentation performance.

\textbf{Implementation Details.}
We implemented CDGC-Net using the PyTorch-based MONAI framework. All experiments were conducted on a single NVIDIA GeForce RTX 2080 Ti GPU with 11~GB of memory. The model was trained for 1,000 epochs using the SGD optimizer with an initial learning rate of 0.01, a momentum of 0.99, a weight decay of $3\times10^{-5}$, and a batch size of 2. For all datasets, we used the same input sizes and preprocessing strategies as the compared methods, without using additional training data. Sliding window inference with 50\% overlap was adopted during testing, and all reported results were obtained from a single model without ensemble strategies. 
For Synapse, the input size was set to $128 \times 128 \times 64$. For ACDC and LA, the input size was set to $160 \times 160 \times 16$. For BraTS, the input size was set to $128 \times 128 \times 128$. Other training hyperparameters and data augmentation settings followed nnFormer~\cite{RN7}.
\begin{table}[!htbp]
\centering
\vspace{-4mm}
\caption{Comparison of segmentation performance on the Synapse dataset in terms of DSC (\%) and HD95 (mm). Spl: spleen; RKid: right kidney; LKid: left kidney; Gal: gallbladder; Liv: liver; Sto: stomach; Aor: aorta; Pan: pancreas.}
\label{tab1}
\scriptsize
\renewcommand{\arraystretch}{1.08}
\setlength{\tabcolsep}{3pt}
\resizebox{\linewidth}{!}{%
\begin{tabular}{@{}lcccccccccc@{}}
\toprule
Method & Spl & RKid & LKid & Gal & Liv & Sto & Aor & Pan & Mean & HD95 \\
\midrule
U-Net~\cite{RN16} & 86.67 & 68.60 & 77.77 & 69.72 & 93.43 & 75.58 & 89.07 & 53.98 & 76.85 & 39.70 \\
TransUNet~\cite{RN6} & 85.08 & 77.02 & 81.87 & 63.16 & 94.08 & 75.62 & 87.23 & 55.86 & 77.49 & 31.69 \\
Swin-Unet~\cite{RN25} & 90.66 & 79.61 & 83.23 & 66.53 & 94.29 & 76.60 & 85.47 & 56.58 & 79.13 & 21.55 \\
Swin UNETR~\cite{RN38} & \textbf{94.59} & 85.88 & 86.51 & 66.72 & 95.33 & 78.20 & 90.75 & 70.07 & 83.51 & 14.78 \\
UNETR~\cite{RN1} & 87.81 & 84.80 & 85.66 & 60.56 & 94.46 & 73.99 & 89.99 & 59.25 & 79.56 & 22.97 \\
CoTr~\cite{RN26} & 88.58 & 83.62 & 85.45 & 68.93 & 93.89 & 76.23 & 85.42 & 63.77 & 80.78 & 19.15 \\
nnFormer~\cite{RN7} & 90.51 & 86.25 & 86.57 & 70.17 & \textbf{96.84} & \textbf{86.83} & 92.04 & \textbf{83.35} & 86.57 & 10.63 \\
UNETR++~\cite{RN15} & 89.67 & 87.23 & 86.04 & \textbf{71.94} & 96.56 & 84.48 & 92.58 & 82.30 & 86.35 & 10.43 \\
nnU-Net~\cite{RN2} & 91.16 & 86.21 & 86.92 & 69.77 & 96.49 & 85.92 & 91.78 & 83.23 & 86.44 & 10.91 \\
CDGC-Net & 91.90 & \textbf{87.58} & \textbf{87.56} & 71.07 & 96.82 & 85.57 & \textbf{92.83} & 82.32 & \textbf{86.96} & \textbf{8.33} \\
\bottomrule
\end{tabular}%
}
\vspace{-4mm}
\end{table}
\subsection{Comparison with State-of-the-Art Methods}
\begin{figure}
    \centering
    \vspace{-3mm}
    \includegraphics[width=1\linewidth]{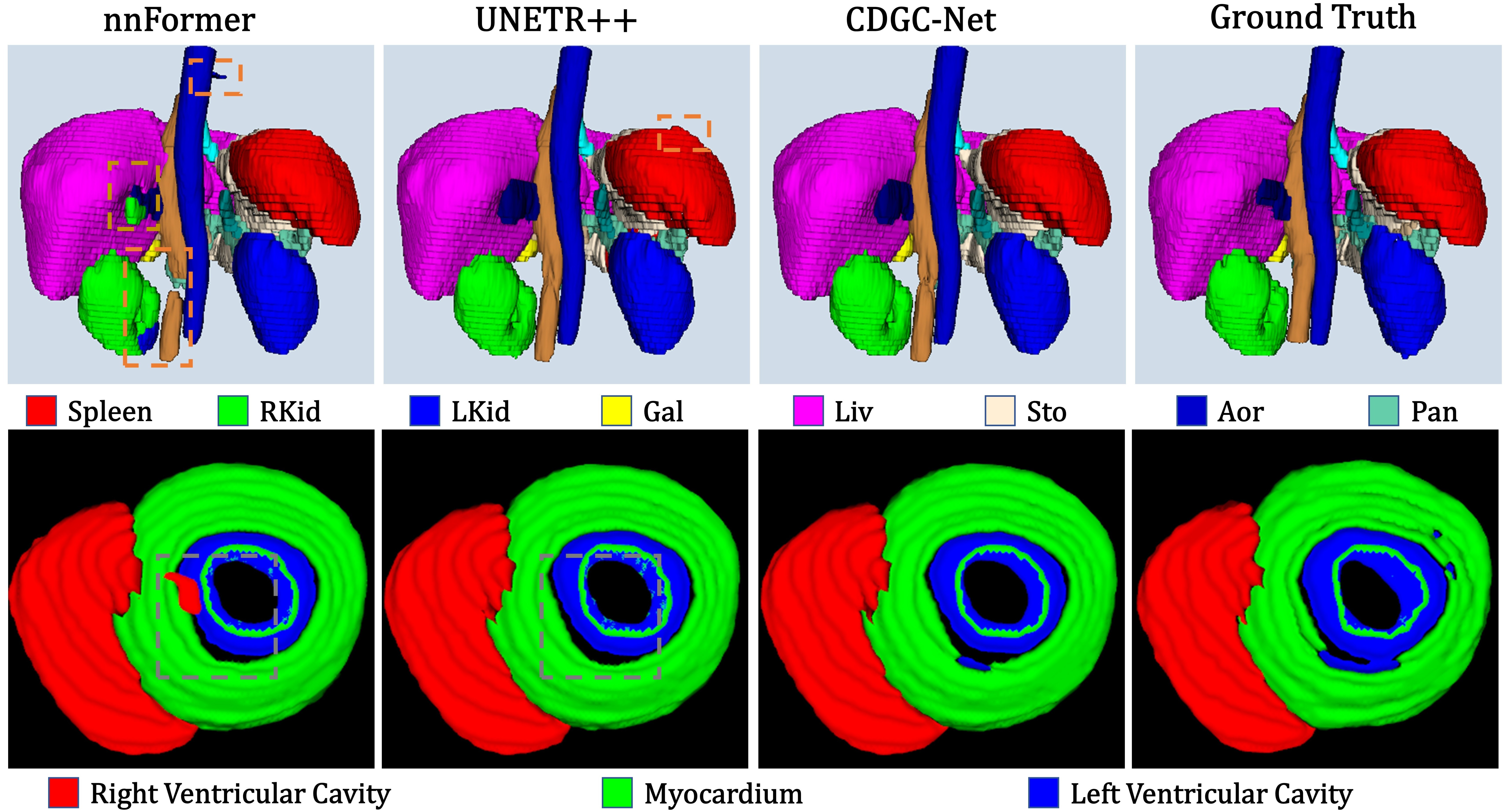}
    \caption{Qualitative comparison of 3D segmentation results on the Synapse (top) and ACDC (bottom) datasets. Orange dashed boxes highlight local segmentation errors produced by the comparison methods.}
    \label{fig2}
    \vspace{-2mm}
\end{figure}

Table~\ref{tab1} reports the segmentation results on the Synapse dataset. CDGC-Net achieved the highest mean DSC of 86.96\% and the lowest HD95 of 8.33~mm among the compared methods. Compared with nnFormer, which achieved the next-highest mean DSC, CDGC-Net improved the mean DSC by 0.39 percentage points and reduced HD95 from 10.63 to 8.33~mm. Relative to UNETR++, CDGC-Net improved the mean DSC by 0.61 percentage points and reduced HD95 from 10.43 to 8.33~mm. CDGC-Net also achieved the highest DSC values for the right kidney, left kidney, and aorta, exceeding the corresponding next-highest values by 0.35, 0.64, and 0.25 percentage points, respectively. In Fig.~\ref{fig2}, the Synapse 3D visualization further shows that CDGC-Net produces more complete organ shapes and clearer boundaries, while the compared methods show local misclassification and boundary errors.

On the ACDC dataset, CDGC-Net achieved the highest DSC among the listed
methods for all three cardiac structures, with a mean DSC of 92.91\%
(Table~\ref{tab2}). Its mean DSC exceeded those of UNETR++, nnU-Net,
and nnFormer by 0.47, 0.50, and 0.85 percentage points, respectively.
The improvement over UNETR++ was statistically significant based on
paired per-case mean DSC values (two-sided Wilcoxon signed-rank test, $p<0.03$). At the structure level, the largest margin over the
next-best result was 0.90 percentage points for the RV. The representative
example in Fig.~\ref{fig2} illustrates that CDGC-Net better preserved the
myocardial ring and reduced local boundary irregularities.
For the BraTS dataset, CDGC-Net obtained the highest mean DSC of 82.56\% and the lowest HD95 of 5.73~mm (Table~\ref{tab3}). Its mean DSC exceeded those of Swin UNETR and UNETR++ by 0.17 and 0.53 percentage points, respectively, while reducing HD95 from 5.92 to 5.73~mm relative to UNETR++. CDGC-Net also achieved the highest DSC for WT and ET, with improvements of 0.14 and 0.20 percentage points, but its TC result was 0.17 percentage points below that of Swin UNETR. In the displayed example in Fig.~\ref{fig3}, its predicted tumor extent and subregions more closely matched the ground truth. 
On the LA dataset, CDGC-Net achieved the highest DSC of 93.52\%, exceeding UNETR++ by 0.32 percentage points (Table~\ref{tab4}). Its HD95 of 3.92~mm was slightly lower than that of UNETR++. The LA example in Fig.~\ref{fig3} shows that CDGC-Net produced a more continuous atrial region and reduced the fragmented predictions highlighted for UNETR++.

\begin{table*}[t]
    \centering

    \begin{minipage}[t]{0.46\textwidth}
        \centering
        \caption{Segmentation performance on the ACDC dataset in terms of DSC (\%).}
        \label{tab2}
        \resizebox{\linewidth}{!}{%
            \begin{tabular}{lcccc}
            \hline
            Method & RV & MYO & LV & Mean \\
            \hline
            TransUNet~\cite{RN6} & 88.86 & 84.54 & 95.73 & 89.71 \\
            Swin-Unet~\cite{RN25} & 88.55 & 85.62 & 95.83 & 90.00 \\
            UNETR~\cite{RN1} & 85.29 & 86.52 & 94.02 & 88.61 \\
            CoTr~\cite{RN26} & 89.13 & 88.40 & 95.16 & 91.04 \\
            nnFormer~\cite{RN7} & 90.94 & 89.58 & 95.65 & 92.06 \\
            UNETR++~\cite{RN15} & 91.08 & 90.28 & 95.97 & 92.44 \\
            nnU-Net~\cite{RN2} & 90.96 & 90.34 & 95.92 & 92.41 \\
            CDGC-Net & \textbf{91.98} & \textbf{90.66} & \textbf{96.09} & \textbf{92.91} \\
            \hline
            \end{tabular}%
        }
    \end{minipage}\hfill
    \begin{minipage}[t]{0.53\textwidth}
        \centering
        \caption{Segmentation performance on the BraTS dataset in terms of DSC (\%) and HD95 (mm).}
        \label{tab3}
        \renewcommand{\arraystretch}{1.015}
        \resizebox{\linewidth}{!}{%
            \begin{tabular}{lccccc}
            \hline
            Method & WT & ET & TC & Mean & HD95 \\
            \hline
            TransUNet~\cite{RN6} & 70.60 & 54.20 & 68.40 & 64.40 & 12.98 \\
            Swin UNETR~\cite{RN38} & 91.12 & 77.65 & \textbf{78.41} & 82.39 & 6.43 \\
            UNETR~\cite{RN1} & 90.35 & 76.30 & 77.02 & 81.22 & 8.82 \\
            CoTr~\cite{RN26} & 91.01 & 77.52 & 77.43 & 81.99 & 9.70 \\
            nnFormer~\cite{RN7} & 91.23 & 77.84 & 77.91 & 82.32 & 6.12 \\
            UNETR++~\cite{RN15} & 91.00 & 77.73 & 77.35 & 82.03 & 5.92 \\
            TransBTS~\cite{RN36} & 90.91 & 77.86 & 76.10 & 81.62 & 9.65 \\
            CDGC-Net & \textbf{91.37} & \textbf{78.06} & 78.24 & \textbf{82.56} & \textbf{5.73} \\
            \hline
            \end{tabular}%
        }
    \end{minipage}
\end{table*}

\begin{figure*}[t]
    \centering
    \begin{minipage}[t]{0.52\textwidth}
        \vspace{6pt} 
        \centering
        \includegraphics[width=\linewidth]{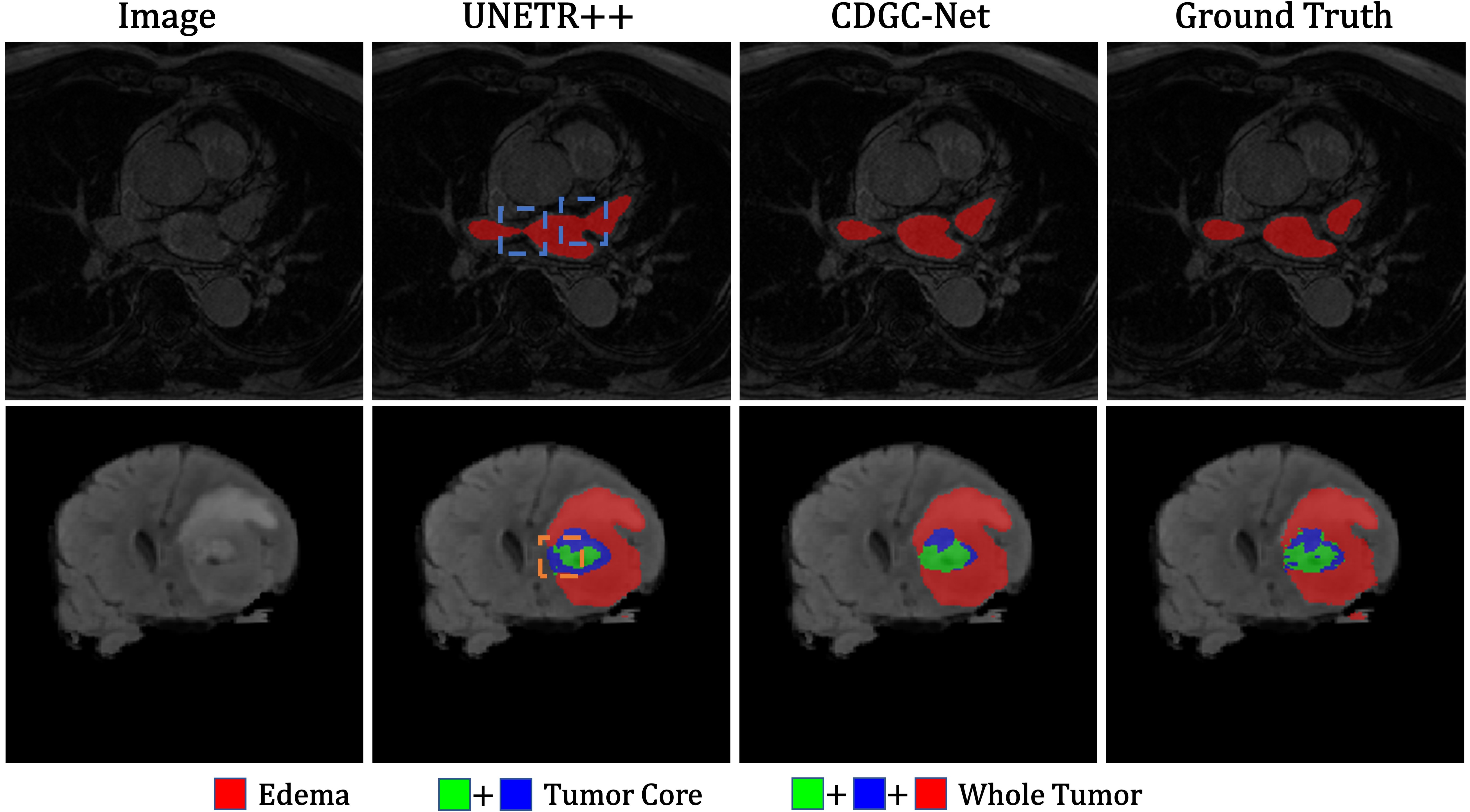}
        \caption{Qualitative comparison on the LA (top) and BraTS (bottom) datasets.}
        \label{fig3}
    \end{minipage}
    \hfill
    \begin{minipage}[t]{0.46\textwidth}
        \vspace{6pt} 
        \centering
        \captionof{table}{Segmentation performance on the LA dataset in terms of DSC (\%) and HD95 (mm).}
        \label{tab4}      
        \renewcommand{\arraystretch}{1.05}
        \setlength{\tabcolsep}{6pt}
        \begin{tabular*}{\linewidth}{@{\extracolsep{\fill}}lcc@{}}
            \hline
            Method & DSC & HD95 \\
            \hline
            V-Net~\cite{RN45}      & 91.14 & 5.75 \\
            DPBNet~\cite{RN47}     & 92.57 & \textbf{2.74} \\
            UNETR++~\cite{RN15}    & 93.20 & 3.95 \\
            nnFormer~\cite{RN7}    & 90.60 & 7.10 \\
            Swin UNETR~\cite{RN38} & 91.40 & 6.20 \\
            CDGC-Net               & \textbf{93.52} & 3.92 \\
            \hline
        \end{tabular*}
    \end{minipage}
    \vspace{-3.5mm}
\end{figure*}
\subsection{Ablation Experiments}
We performed ablation experiments on ACDC to quantify the contribution of each CDGC component. As reported in Table~\ref{tab:ablation}, the complete CDGC-Net achieved a mean DSC of 92.91\%, improving upon the UNETR++ baseline of 92.44\% by 0.47 percentage points.
Among the evaluated components, excluding the global branch caused the largest decrease, lowering the mean DSC by 0.31 percentage points. Removing residual feature alignment produced a decrease of 0.25 percentage points. The window branch and GHCA contributed improvements of 0.18 and 0.19 percentage points, respectively. Replacing the shared key with separate key projections decreased the mean DSC from 92.91\% to 92.77\%, indicating a contribution of 0.14 percentage points from key sharing. Table~\ref{tab:complexity} summarizes the model complexity for an input size of $64\times128\times128$. CDGC-Net contains 25.83M parameters and requires 28.62G FLOPs. Relative to UNETR++, these values correspond to reductions of 39.87\% and 40.30\%, respectively.

\begin{table*}[t]
\centering
\begin{minipage}[t]{0.56\textwidth}
\centering
\captionsetup{font=footnotesize}
\caption{Ablation results of different components in the CDGC block on the ACDC dataset.}
\label{tab:ablation}
\footnotesize
\renewcommand{\arraystretch}{1.08}
\setlength{\tabcolsep}{6pt}
\resizebox{\linewidth}{!}{%
\begin{tabular}{lcccc}
\toprule
Ablation setting & RV & MYO & LV & Mean \\
\midrule
Baseline & 91.08 & 90.28 & 95.97 & 92.44 \\
w/o Global branch & 91.34 & 90.48 & 95.97 & 92.60 \\
w/o Window branch & 91.65 & 90.64 & 95.92 & 92.73 \\
w/o GHCA & 91.55 & 90.59 & 96.02 & 92.72 \\
w/o Key sharing & 91.69 & 90.61 & 96.01 & 92.77 \\
w/o RFA & 91.37 & 90.55 & 96.05 & 92.66 \\
CDGC-Net & \textbf{91.98} & \textbf{90.66} & \textbf{96.09} & \textbf{92.91} \\
\bottomrule
\end{tabular}%
}
\end{minipage}
\hfill
\begin{minipage}[t]{0.40\textwidth}
\centering
\captionsetup{font=footnotesize}
\caption{Model complexity comparison under the same input size.}
\label{tab:complexity}
\footnotesize
\renewcommand{\arraystretch}{1.515}
\setlength{\tabcolsep}{7pt}
\resizebox{\linewidth}{!}{%
\begin{tabular}{lcc}
\toprule
Method & Params (M) & FLOPs (G) \\
\midrule
UNETR~\cite{RN1} & 115.38 & 586.12 \\
TransUNet~\cite{RN6} & 96.07 & 97.36 \\
Swin UNETR~\cite{RN38} & 62.19 & 383.57 \\
nnFormer~\cite{RN7} & 150.05 & 213.37 \\
UNETR++~\cite{RN15} & 42.96 & 47.94 \\
CDGC-Net & \textbf{25.83} & \textbf{28.62} \\
\bottomrule
\end{tabular}%
}
\end{minipage}
\end{table*}
\section{Conclusion}
In this paper, we propose CDGC-Net, an efficient 3D medical image segmentation network based on cooperative spatial-channel feature modeling. The core CDGC block integrates CDSA and GHCA within the same module. CDSA captures complementary global and local spatial dependencies, while GHCA models grouped channel relationships to improve channel discrimination. A shared key $K$ connects the two modules in a unified feature space, allowing spatial information to guide channel recalibration and reducing redundant projections. Residual feature alignment further stabilizes the feature refinement process and preserves the original representation.
Experiments on Synapse, ACDC, BraTS, and LA show that CDGC-Net achieves competitive or superior Dice and HD95 performance across multi-organ, cardiac, brain tumor, and left atrium segmentation tasks. It also uses fewer parameters and lower FLOPs, demonstrating a better balance between segmentation accuracy and computational efficiency. In future work, we will further evaluate CDGC-Net under cross-domain and multi-center settings and extend it to more heterogeneous medical imaging scenarios.

\begin{credits}
\subsubsection{\ackname} This work was supported by the National Natural Science Foundation of China under Grant 62261039 and the Postgraduate Innovation Special Fund of Nanchang Hangkong University under Grant YC2025-060.

\subsubsection{\discintname}
The authors have no competing interests to declare that are relevant to the content of this article.
\end{credits}

\bibliographystyle{splncs04}
\bibliography{reference}

\end{document}